\documentclass{article}

\usepackage[square,numbers]{natbib}
\usepackage[preprint]{neurips_2021}

\usepackage[utf8]{inputenc} 
\usepackage[T1]{fontenc}    
\usepackage{hyperref}       
\usepackage{url}            
\usepackage{booktabs}       
\usepackage{amsfonts}       
\usepackage{nicefrac}       
\usepackage{microtype}      
\usepackage{color}         
\usepackage[dvipsnames]{xcolor}

\usepackage{pifont}
\newcommand{\cmark}{\ding{51}}%
\newcommand{\xmark}{\ding{55}}%

\usepackage{times}
\usepackage{epsfig}
\usepackage{graphicx}
\usepackage{amsmath}
\usepackage{amssymb}
\usepackage{mathrsfs}
\usepackage{tabularx}
\usepackage[para,online,flushleft]{threeparttable}
\usepackage{multirow}
\usepackage{subcaption}
\usepackage[ruled,linesnumbered]{algorithm2e}
\usepackage[title]{appendix}
\usepackage{caption}
\usepackage{float}
\usepackage{amsmath}

\usepackage{enumitem}
\usepackage{wrapfig}
\usepackage{comment}

\newfloat{figtab}{htb}{fgtb}
\makeatletter
  \newcommand\figcaption{\def\@captype{figure}\caption}
  \newcommand\tabcaption{\def\@captype{table}\caption}
\makeatother

\def\eg{\emph{e.g.}} 
\def\ie{\emph{i.e.}}

\def\etal{\emph{et al.}}

\colorlet{dark-blue}{blue!65!black}
\colorlet{dark-green}{green!55!black}
\colorlet{dark-red}{red!80!black}
\colorlet{dark-yellow}{yellow!90!black}
\colorlet{white-blue}{blue!70!green}
\definecolor{mypink}{RGB}{219, 48, 122}

\hypersetup{
    colorlinks=true,%
    citecolor=dark-blue,%
    filecolor=dark-blue,%
    linkcolor=red,%
    urlcolor=magenta
}

\title{Source-Free Open Compound Domain Adaptation in Semantic Segmentation}

\author{%
  Yuyang Zhao$^{\textcolor{mypink}{1}}$\thanks{Equal contribution.} \quad Zhun Zhong$^{\textcolor{mypink}{2}*}$ \quad Zhiming Luo$^{\textcolor{mypink}{3}}$ \quad Gim Hee Lee$^{\textcolor{mypink}{1}}$ \quad Nicu Sebe$^{\textcolor{mypink}{2}}$\\
  {$^{\textcolor{mypink}{1}}$ National University of Singapore} \quad
  {$^{\textcolor{mypink}{2}}$ University of Trento} \quad 
  {$^{\textcolor{mypink}{3}}$ Xiamen University} \\

}

\begin{document}

\maketitle

\begin{abstract}

In this work, we introduce a new concept, named source-free open compound domain adaptation (SF-OCDA), and study it in semantic segmentation. SF-OCDA is more challenging than the traditional domain adaptation but it is more practical. It jointly considers (1) the issues of data privacy and data storage and (2) the scenario of multiple target domains and unseen open domains. In SF-OCDA, only the source pre-trained model and the target data are available to learn the target model. The model is evaluated on the samples from the target and unseen open domains.
To solve this problem, we present an effective framework by separating the training process into two stages: (1) pre-training a generalized source model and (2) adapting a target model with self-supervised learning. In our framework, we propose the Cross-Patch Style Swap~(CPSS) to diversify samples with various patch styles in the feature-level, which can benefit the training of both stages. First, CPSS can significantly improve the generalization ability of the source model, providing more accurate pseudo-labels for the latter stage. Second, CPSS can reduce the influence of noisy pseudo-labels and also avoid the model overfitting to the target domain during self-supervised learning, consistently boosting the performance on the target and open domains. Experiments demonstrate that our method produces state-of-the-art results on the C-Driving dataset. Furthermore, our model also achieves the leading performance on CityScapes for domain generalization.

\end{abstract}

\section{Introduction}

Deep learning has now achieved a remarkable success in fully-supervised semantic segmentation~\cite{fcn,zhao2017pyramid, deeplab}, which, however, is relied heavily on the expensive dense pixel-wise annotations. One solution to lighten the labeling cost is unsupervised domain adaptation~(UDA), which aims to transfer the knowledge of labeled synthetic data to unlabeled real-world data. Despite the effectiveness of existing UDA methods~\cite{adaptseg,zhang2021prototypical, zheng2021rectifying}, they mainly consider the context of a single target domain, resulting in limited applications in the real-world. Indeed, the target domain may be captured from multiple data distributions without a clear separation and the system will unavoidably face instances from unseen domains. To investigate a more realistic domain adaptation problem, in this paper, we consider the setting of open compound domain adaptation~(OCDA)~\cite{liu2020open} for semantic segmentation. In OCDA, the unlabeled target domain is a compound of multiple homogeneous domains without domain labels. The adapted model is applied to test samples from the compound target domain and an open domain, where the open domain is unseen during training.

\begin{figure}[t]
    \centering
    \includegraphics[width=0.97\linewidth]{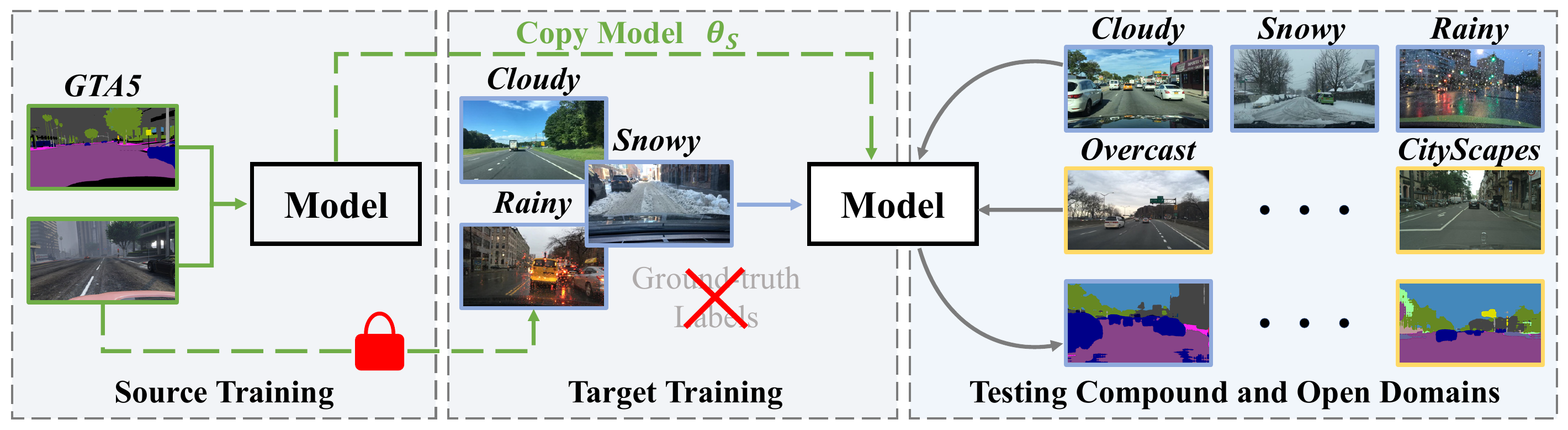}
    \vspace{-.09in}
    \caption{Illustration of source-free open compound domain adaptation (SF-OCDA). In the training stage, the model is first trained on the (synthetic) labeled \textcolor{ForestGreen}{source} data and then adapted to the (real-world) unlabeled \textcolor{Cerulean}{compound} target data. The \textcolor{ForestGreen}{source} data are \textbf{not available} during the target adaptation. In the testing stage, the learned model is used to predict the semantic segmentation results for samples from the \textcolor{Cerulean}{compound} and \textcolor{dark-yellow}{open} domains.}
    \label{fig:intro}
    \vspace{-.1in}
\end{figure}

Existing UDA~\cite{adaptseg,zhang2021prototypical,advent} and OCDA~\cite{liu2020open,park2020discover,gong2020cluster} methods commonly require the use of the labeled source data during the whole training process. However, the source data are not always available due to data privacy. In addition, the source data are generally very large, 
\begin{wraptable}{r}{0.51\textwidth}
    \caption{Comparisons of different adaptation settings. \textbf{DA}: domain adaptation, \textbf{SF-DA}: source-free DA, \textbf{DG}: domain generalization, \textbf{OCDA}: open compound DA, \textbf{SF-OCDA}: source-free OCDA.}
    \label{tab:settings}
    \scriptsize
    \centering
    \begin{tabularx}{0.51\textwidth}{p{1.2cm}|m{0.55cm}<{\centering}m{0.55cm}<{\centering}m{0.85cm}<{\centering}m{0.75cm}<{\centering}m{0.55cm}<{\centering}}
    \toprule
    \multirow{2}{*}{Settings} & \multicolumn{1}{c}{Source} & Source & Unlabeled & \multicolumn{1}{c}{Multiple} & Open \\
    ~ & Data & Model & Target & Targets & Targets \\
    \midrule
    DA~\cite{adaptseg} & \color{ForestGreen}{\cmark} & \color{ForestGreen}\cmark & \color{ForestGreen}\cmark & \color{red}\xmark & \color{red}\xmark  \\
    SF-DA~\cite{liu2021source}	& \color{red}\xmark &\color{ForestGreen}\cmark & \color{ForestGreen}\cmark & \color{red}\xmark &  \color{red}\xmark \\
    DG~\cite{DRPC} & \color{red}\xmark & \color{ForestGreen}\cmark & \color{red}\xmark & \color{red}\xmark & \color{ForestGreen}\cmark \\
    OCDA~\cite{liu2020open} & \color{ForestGreen}\cmark & \color{ForestGreen}\cmark & \color{ForestGreen}\cmark & \color{ForestGreen}\cmark & \color{ForestGreen}\cmark \\
    SF-OCDA & \color{red}\xmark & \color{ForestGreen}\cmark & \color{ForestGreen}\cmark & \color{ForestGreen}\cmark & \color{ForestGreen}\cmark\\
    \bottomrule
    \end{tabularx}
\end{wraptable}
which require plenty of storage space (\eg, GTA5~\cite{gta5}$\approx$57GB). This further limits the applications of existing methods, especially when transferring to a lightweight self-driving device.
 Nevertheless, we can choose to maintain the pre-trained source model instead of the source data, enabling us to obey the data privacy policy and use much less storage space (\eg, DeepLab-VGG16~\cite{deeplab,vgg}$\approx$120MB). These facts motivate us to introduce a more challenging but practical setting for OCDA, called source-free OCDA (\textbf{SF-OCDA}), where only the source pre-trained model and the unlabeled target data are available during the training of the target model. In the literature, source-free domain adaptation~(SF-DA) has recently been developed in image classification~\cite{liang2020we,kundu2020universal} and semantic segmentation~\cite{liu2021source} for the single target case. However, as shown in Fig.~\ref{fig:intro} and Tab.~\ref{tab:settings}, compared with SF-DA, our SF-OCDA demands not only adapting to data from multiple target domains but also considering the generalization performance on unseen domains.

In SF-OCDA, the source data and target data are invisible to each other. In such context, we cannot align the domain distributions as traditional DA methods~\cite{adaptseg,zhang2021prototypical, zheng2021rectifying}. Instead, this paper introduces an effective two-stage framework for SF-OCDA, which consists of (1) training a generalized source model and (2) adapting the target model with self-supervised learning. In the first stage, we aim to learn a robust model, which can generalize well to different target domains. To achieve this goal, we propose the \textbf{C}ross-\textbf{P}atch \textbf{S}tyle \textbf{S}wap~(CPSS), which can effectively augment the samples with various image styles. Specifically, CPSS first extracts the styles of patches in feature maps and then randomly exchanges the styles among patches by the instance normalization and de-normalization. In this manner, CPSS can prevent the model from overfitting to the source domain and thus significantly improve the generalization ability of the model. In the second stage, we adapt the target model by self-supervised learning. 
Specifically, we optimize the target model with the guide of pseudo-labels generated from the pre-trained source model, which can implicitly align the source and target distributions under the constraint of label consistency. 
Moreover, CPSS is also applied to reduce the influence of noisy pseudo-labels and to avoid overfitting to the target domain, which can further boost the performance on the compound and open domains. Our contributions are summarized as follows:
\begin{itemize}[leftmargin=0.15in]
\setlength{\itemsep}{0pt}
\setlength{\parsep}{0pt}
\item We introduce a new setting for semantic segmentation, \textit{i.e.}, source-free open compound domain adaptation (SF-OCDA), which is an important yet unstudied problem. In addition, we propose an effective framework for solving SF-OCDA, which focuses on learning a generalized model during the stages of source pre-training and target adaptation.

\item We propose the CPSS, which diversifies the samples in the feature-level, to improve the generalization ability of the model in both source and target training stages. CPSS is a lightweight module without learnable parameters, which can be readily injected into existing segmentation models.

\item The proposed framework learned with the source-free constraint significantly outperforms the state-of-the-art methods on the OCDA benchmark. Our approach also surpasses the advanced domain generalization approaches on CityScapes. 
	
\end{itemize}

\section{Related Work}
\textbf{Transfer Learning in Semantic Segmentation}. 
To tackle the expensive cost of collecting and labeling real-world data, transfer learning has attracted a widespread attention in semantic segmentation. Commonly, transfer learning methods are developed along two directions: unsupervised domain adaptation (UDA) and domain generalization (DG). UDA aims at transferring the knowledge from a labeled source domain to an unlabeled target domain. Existing UDA approaches can be roughly divided into two categories, \ie, aligning domain distributions through adversarial learning~\cite{adaptseg,advent,luo2019taking,du2019ssf} and self-training on the target domain~\cite{cbst,zhang2021prototypical,ma2021coarse,zheng2021rectifying}.
DG focuses on training a robust model with synthetic data, which can generalize well on unseen real-world target data. To reduce the large gap between synthetic data and the real-world data, DG methods usually augment the synthetic samples~\cite{DRPC,FSDR} with the styles of ImageNet~\cite{imagenet} or conditionally align the outputs~\cite{ASG,CSG} between the segmentation model and the ImageNet pre-trained model. On the other hand, some works are proposed to learn domain-invariant features by removing domain-specific information~\cite{robustnet,ibn} or feature augmentation~\cite{crossnorm}.
Recently, Liu \etal~\cite{liu2020open} propose the setting of open compound domain adaptation (OCDA), which can be regarded as an extension of UDA and DG. In OCDA, the model trained with source and target data is used to evaluate samples from the compound target domain and unseen open domain. Liu \etal~\cite{liu2020open} introduce a memory-based curriculum learning framework to improve the generalization on the compound and open domains. Park \etal~\cite{park2020discover} and Gong \etal~\cite{gong2020cluster} discover the latent target domains and align the source and latent domains with multiple domain discriminators. Different from these OCDA methods, this work investigates the OCDA under the source-free constraint and aims to learn a robust model by augmenting features with patch styles.

\textbf{Source-Free Domain Adaptation}.
Hypothesis transfer learning (HTL)~\cite{kuzborskij2013stability} aims to retain the prior knowledge in a form of hypothesis instead of data for the source domain. However, the main drawback of HTL is that it requires a small set of labeled target data. Inspired by HTL, source-free domain adaptation (SF-DA)~\cite{chidlovskii2016domain, liang2019distant,liu2021source,liang2020source,liang2020we,kundu2020universal,tian2021vdm, hou2021visualizing} has recently flourished in the domain adaptation community. In SF-DA, instead of the source data, the source pre-trained model is provided in the target training stage.
SHOT~\cite{liang2020we} maintains the source hypothesis by fixing the trained classifier and maximizes mutual information of target outputs for distribution alignment. Kundu \etal~\cite{kundu2020universal} generate negative samples by image composition, which are used to narrow domain shift and category gap during source training. In addition, an instance-level weighting mechanism is proposed for effective target adaptation. Lately, Liu \etal~\cite{liu2021source} introduce source-free domain adaptation for semantic segmentation and utilize the batch normalization statistics of the source model to recover source-like samples. In this work, we introduce the source-free open compound domain adaptation~(SF-OCDA) for semantic segmentation, extending SF-DA to a more realistic setting. The comparisons between SF-OCDA and existing adaptation settings are reported in Tab.~\ref{tab:settings}.

\section{Methodology}

\begin{figure}[ht]
    \centering
    \includegraphics[width=0.95\linewidth]{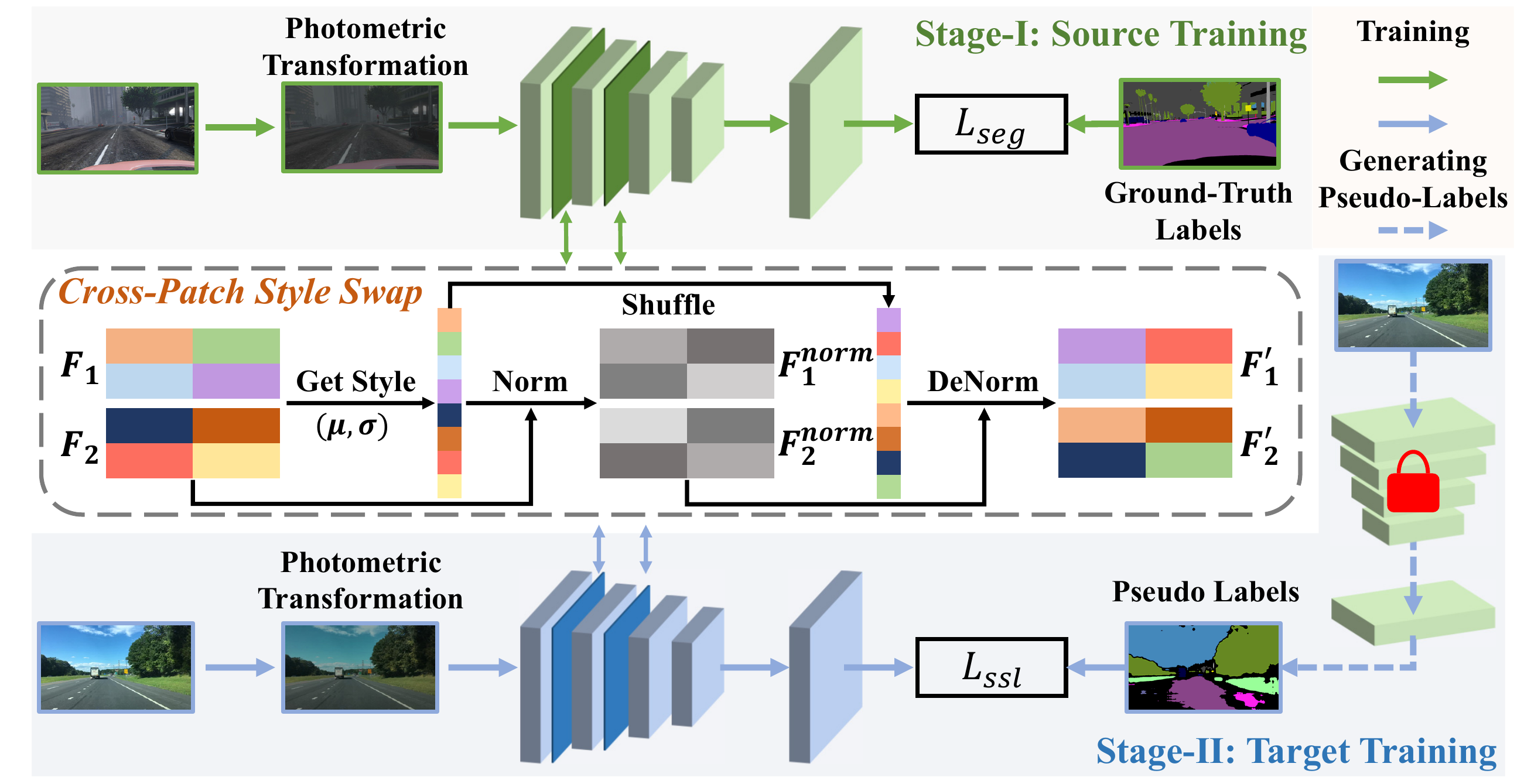}
    \vspace{-.11in}
    \caption{The framework of the proposed method. (1) The model is first trained on the labeled source domain. (2) We generate pseudo-labels by the source pre-trained model and train the target model in a self-training manner. In the second stage, we have no access to the source data. To improve the generalization ability of the model, we equip the model with the Cross-Patch Style Swap module in the two training stages, which augments features by exchanging styles among patches.}
    \label{fig:framework}
\end{figure}

\subsection{Preliminaries}

In open compound domain adaptation~(OCDA)~\cite{liu2020open}, we are given a  labeled (synthetic) source domain $\mathcal{S}$ and an unlabeled (real) compound target domain $\mathcal{T}$. The goal is to train a model that can accurately predict semantic labels for instances from the compound and open target domains. Specifically, $\mathcal{S}$ includes $N_S$ images $x^s_i\in \mathbb{R}^{H^s\times W^s \times 3}$ and their corresponding semantic labels $y^s_i\in \mathbb{R}^{H^s\times W^s}$ of $C$ classes.  $\mathcal{T}$ contains $N_T$ images $x^t_i\in \mathbb{R}^{H^t\times W^t \times 3}$ of multiple homogeneous domains without semantic and domain labels. In this paper, \textbf{we consider the setting of source-free OCDA (SF-OCDA)}, which imposes an extra constraint that only the pre-trained source model, instead of the source data, is available for training the target model together with the unlabeled target data.

\subsection{Overview}
In this Section, we propose an effective framework for SF-OCDA, which separates the training process into two stages: (1) training a generalized source model and (2) adapting a target model with self-supervised learning. We also introduce the Cross-Patch Style Swap (CPSS) to augment features with various patch styles, which can significantly improve the generalization ability of the model in both training stages. The pipeline of our framework is shown in Fig.~\ref{fig:framework}. Next, we first introduce our CPSS module (Sec.~\ref{sec:cpss}) and then present the proposed training strategy (Sec.~\ref{sec:training}) in detail.

\begin{figure}[t]
    \centering
    \begin{subfigure}[t]{0.42\textwidth}
        \centering
        \includegraphics[width=1\textwidth]{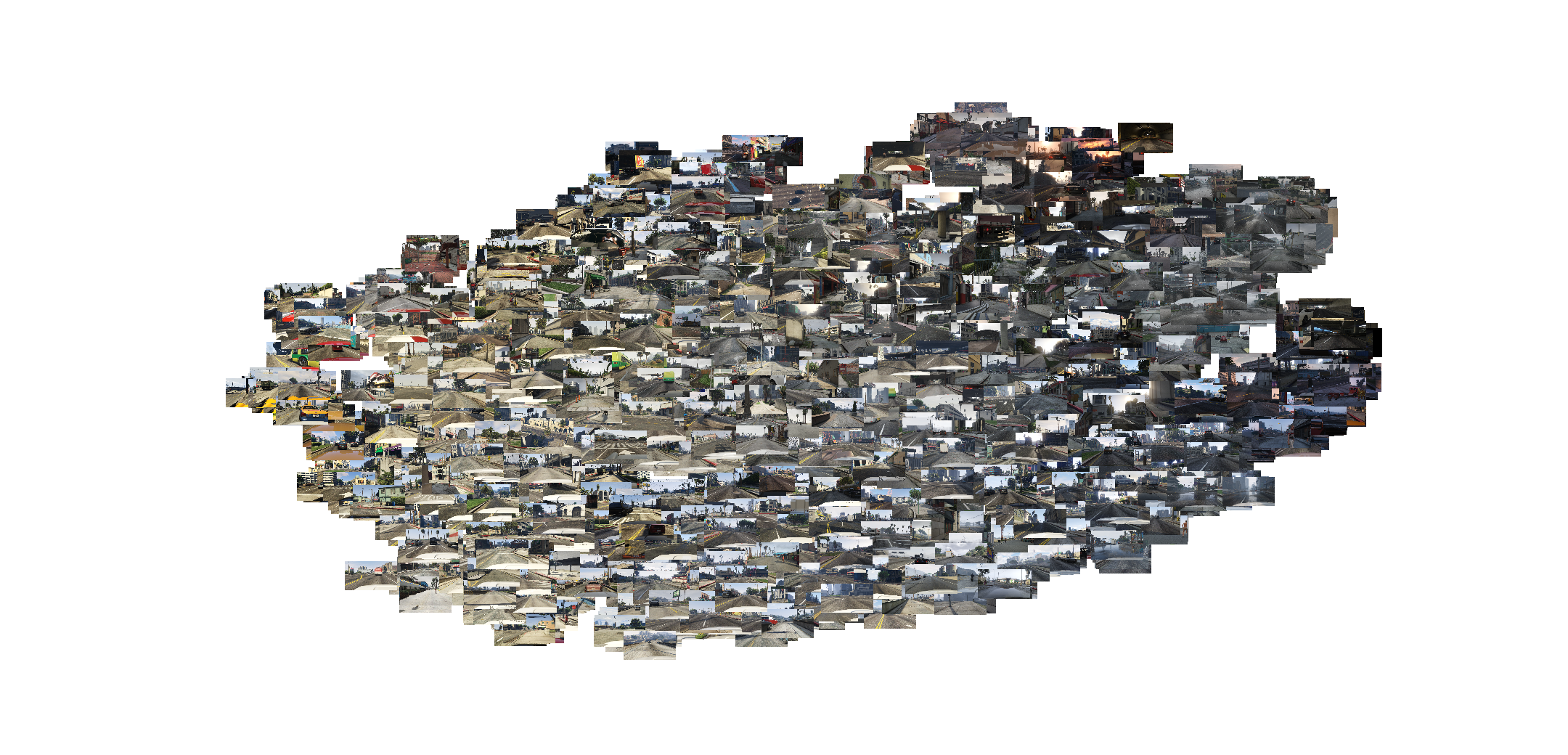}
        \vspace{-.3in}
        \subcaption{Source Domain~(GTA5~\cite{gta5}).}
    \end{subfigure}
    \begin{subfigure}[t]{0.42\textwidth}
        \centering
        \includegraphics[width=1\textwidth]{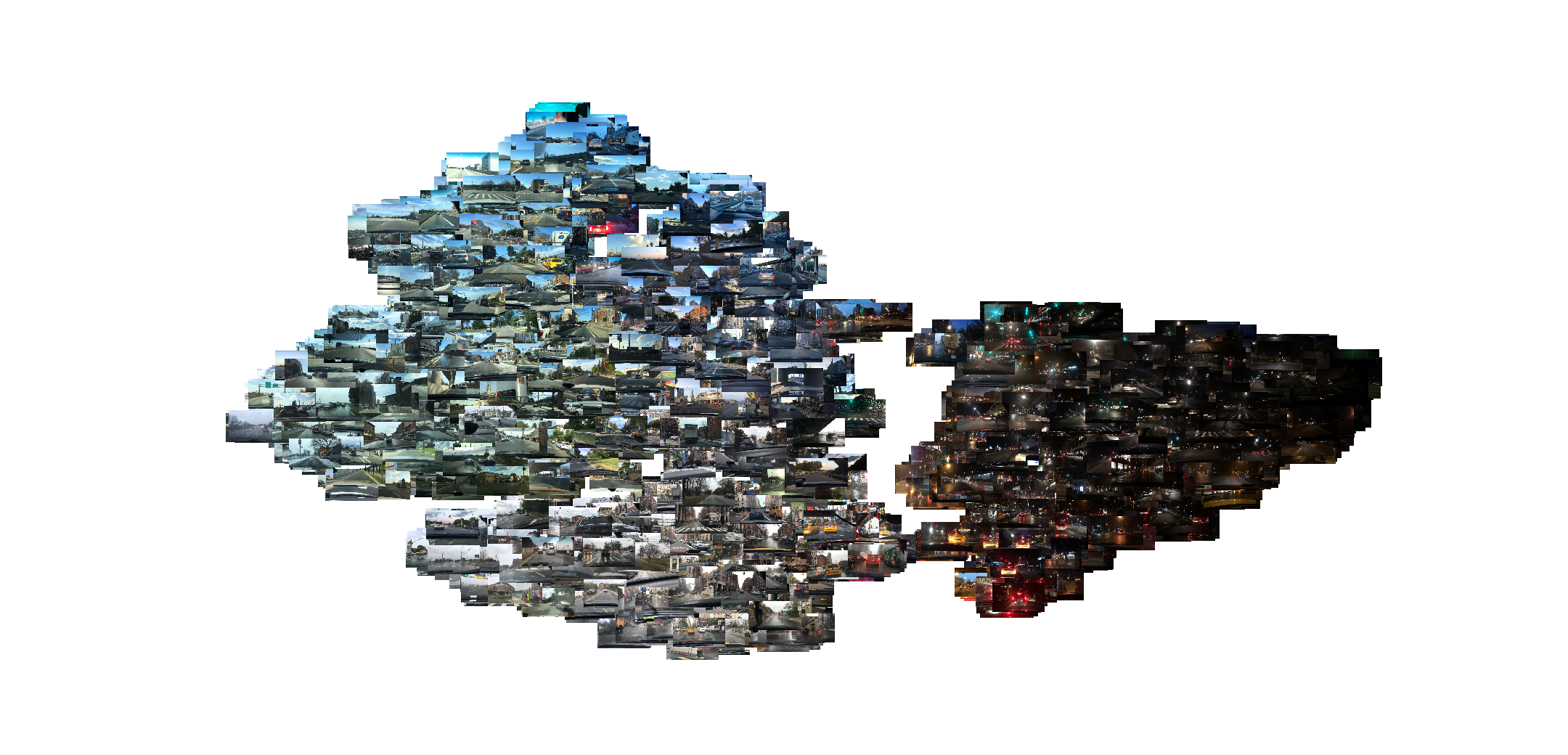}
        \vspace{-.3in}
        \subcaption{Target Domain~(C-Driving~\cite{liu2020open}).}
    \end{subfigure}
    \vspace{-.05in}
    \caption{Visualization of style distributions for (a) source domain  and (b) target domain. We use the concatenation of mean and standard deviation of the feature map after the first block of VGG-16~\cite{vgg} as the style feature and show the 2D embeddings by t-SNE~\cite{tsne}. Zoom in for details.}
    \label{fig:style_tsne}
\end{figure}

\subsection{Cross-Patch Style Swap}
\label{sec:cpss}

\textbf{Motivation}. Image style variation is an important factor that influences the model performance in semantic segmentation. Although the synthetic data are built to simulate the real-world images, the styles of synthetic images are still very different from that of the real ones. Therefore, the model trained on the synthetic data will be sensitive to the real style variations and thus produces poor performance on real images. To this end, we attempt to learn a robust model, which is insensitive to style variations, by augmenting the training samples with diverse styles.

In order to implement style augmentation, the key is extracting style factors from images. To achieve this goal, we draw inspiration from the style transfer~\cite{in, dumoulin2016learned}, which obtains image styles by extracting the mean $\mu$ and standard deviation $\sigma$ of the feature map in a network. In Fig.~\ref{fig:style_tsne}, we visually verify the feasibility of using the $\mu$ and $\sigma$ as the style features in semantic segmentation. It is clear that images of various styles ~(\eg, day and night) can be well-separated by the style features. In addition, AdaIN~\cite{adain} shows that an input sample can be transferred to an arbitrary style while remaining the semantic content, by replacing the style features, formulated as:
\begin{equation}
    \text{AdaIN}(x, y)=\sigma(y)\left(\frac{x-\mu(x)}{\sigma(x)}\right)+\mu(y),
\end{equation}
where $\mu(.)$ and $\sigma(.)$ denote the mean and standard deviation of the input feature map, respectively. $x$ and $y$ are two feature maps that provide the semantic content and the image style, respectively. Inspired by AdaIN, we propose two style augmentation operations based on the style features of image patches for training a robust segmentation model.

\textbf{Intra-Image Cross-Patch Style Swap.} In the self-driving scenario, different patches (\textit{e.g.}, up and down) of a frame may include different objects, such as sky, vehicle, road and fence, making these patches present different styles. Intuitively, we can generate a new stylized sample by exchanging the style features of different patches. Hence, we propose the intra-image Cross-Patch Style Swap. Specifically, the feature map of an image is first separated into $n=n_h\times n_w$ patches:
\begin{equation}
F =
{\left[                
  \begin{array}{ccc}   
    F_{1,1} & \cdots   & F_{1,n_w}   \\
    \vdots & \ddots   & \vdots  \\
    F_{n_h,1}  & \cdots & F_{n_h,n_w}  \\
  \end{array}
\right]}.
\end{equation}

After that, each patch is normalized by the mean and standard deviation of itself, and de-normalized by the feature style of a random patch, formulated by:
\begin{equation}
\label{eq:intra-cpss}
F^{'}_{i,j}=\sigma(\tilde{F}_{i,j})\left(\frac{F_{i,j}-\mu(F_{i,j})}{\sigma(F_{i,j})}\right)+\mu(\tilde{F}_{i,j}),
\end{equation}
where $F^{'}_{i,j}$ denotes the style swapped counterpart of $F_{i,j}$. $\tilde{F}_{i,j}$ denotes the shuffled patch that provides the style feature.

\textbf{Inter-Image Cross-Patch Style Swap.} Although the intra-image CPSS can enrich the styles of a feature map, the model can easily remember the intra-image style variations after several training epochs, which will limit the effectiveness of the CPSS. However, the patch styles vary greatly among different images, which can be used to further enhance the style diversity during CPSS. Taking this into consideration, we introduce the inter-image CPSS, which collects style features from all the patches in a mini-batch with $B$ samples and exchanges these styles~($B\times n$) among all patches. We re-formulate Eq~\ref{eq:intra-cpss} as:
\begin{equation}
F^{'}_{k,i,j}=\sigma(\tilde{F}_{k,i,j})\left(\frac{F_{k,i,j}-\mu(F_{k,i,j})}{\sigma(F_{k,i,j})}\right)+\mu(\tilde{F}_{k,i,j}),
\end{equation}
where $F^{'}_{k, i,j}$ denotes the swapped counterpart of patch $F_{i,j}$ in the $k$th sample. $\tilde{F}_{k,i,j}$ denotes a randomly selected patch that provides the style feature.

CPSS is injected into several layers of the backbone, which is activated in the training stage with a probability of $\beta$ and is not used in the testing stage. 

\textbf{Photometric Transformation}. In practice, the brightness, contrast and saturation of the frame vary in different situations. For example, images are brighter in sunny morning while the contrast is stronger in snowy weather. In addition, there may exist blurry images caused by the rainy weather. Consequently, we randomly apply photometric transformation to the input images, including color jitter, Gaussian blur and grayscale, to simulate the real-world style various, which can further improve the effect of CSPP.

\subsection{Model Training}
\label{sec:training}

As shown in Fig.~\ref{fig:framework}, our framework includes two stages, \ie, the source training stage and the target training stage, where the source data and target data are used independently in their own stages.

\textbf{Stage-I: Source Training.} In this stage, we aim at training a generalized model with synthetic labeled source domain $\mathcal{S}$. We adopt the cross-entropy loss to train the model, formulated as:
\begin{equation}
    L_{seg} = - \sum\limits_{m=1}^{H \times W} \sum\limits_{c=1}^C y^s_{m,c} \log p^s_{m,c},
\end{equation}
where $y^s_{m,c}$ denotes the ground truth for the $m$th pixel and $p^s_{m,c}$ denotes the softmax probability of this pixel belonging to the $c$th class. Importantly, we employ the proposed CPSS along with photometric transformation to augment the samples in both feature- and image-levels, which can effectively improve the generalization ability of the source model.

\textbf{Stage-II: Target Training.} For SF-OCDA, source data are not available in this stage. Instead, we are given the source pre-trained model and the unlabeled compound target domain $\mathcal{T}$ to learn a target model that can perform well on both compound and open domains. 
In this stage, the target model is cloned from the source pre-trained model and trained in a self-supervised manner. 

Specifically, we first generate pseudo-labels based on the predictions of the source pre-trained model by maximum probability threshold (MPT)~\cite{li2019bidirectional}. MPT estimates class thresholds based on the top $q$\% pixels of each class and a predefined threshold $\tau$. The pseudo-labels are then assigned to pixels where the prediction values of the dominant classes are higher than the corresponding class thresholds.

With the pseudo-labels, we employ the cross-entropy loss to enforce the consistency between the source and target outputs:
\begin{equation}
    L_{ssl} = - \sum\limits_{m=1}^{HW} \sum\limits_{c=1}^C \hat{y}^t_{m,c} \log p^t_{m,c},
\end{equation}
where $p^t_{m,c}$ is the prediction of the target model and $\hat{y}^t_{m,c}$ is the generated pseudo-label. Note that, we only update the model with pixels that are assigned with pseudo-labels, and ignore the others. 

Similar to Stage-I, we also adopt CPSS and the photometric transformation to train the target model, which brings two advantages. First, the negative impact of noisy pseudo-labels can be reduced by training on samples with more augmentations~\cite{sohn2020fixmatch}. Second, CPSS can prevent the model from overfitting to the styles of the target domain, leading the model to be more robust to style variations. These two advantages improve the model performance on the target compound and open domains.

\section{Experiments}

\subsection{Experimental Setup}
\label{sec:details}

\textbf{Datasets.} Following \cite{liu2020open}, we use the synthetic image data GTA5~\cite{gta5} as the source domain, the rainy, snowy, and cloudy images in C-Driving~\cite{bdd100k,liu2020open} as the compound target domain, and the overcast images in C-Driving as the open domain. To further measure the generalization ability of models, we additionally use Cityscapes~\cite{cityscapes} as an extended open domain. GTA5 includes 24,966 training images with a resolution of 1914$\times$1052. C-Driving consists of 14,697 unlabeled training images and 1,430 testing images, where the image size is 1280$\times$720. Cityscapes contains 500 images of 2048$\times$1024 for validation. For all datasets, pixels belong to 19 shared semantic categories. During testing, we use mean intersection-over-union~(mIoU) to evaluate the semantic segmentation performance. 


\textbf{Implementation Details}. 
We use the DeepLab-V2~\cite{deeplab} with VGG16~\cite{vgg} backbone as the segmentation model. For the source training stage, following~\cite{adaptseg,park2020discover}, we use SGD with an initial learning rate $2.5\times10^{-4}$, momentum 0.9 and weight decay $5\times10^{-4}$ to optimize the model. For the target training stage, the learning rate is reduced to $1\times10^{-4}$. In both stages, we use the polynomial decay with a power of 0.9 to schedule the learning rate. The total training process takes 150K iterations, with a batch size of 1. We set $\tau$=0.9 and $q$\%=50\% for generating pseudo-labels. For CSPP, the number of patches $n$ and the activation probability $\beta$ are set to 4 and 0.3, respectively. By default, we use the inter-image CSPP and inject it after the first and second blocks of the VGG16. Note that, we use a batch size of 4 for CPSS, but optimize the model with only the first image. This can greatly reduce the computational cost, because using a batch size of 1 or 4 achieves a similar performance. All models are trained with one GTX 2080 TI GPU (11GB).

\subsection{Comparison with State-of-the-Art Methods}

\begin{table}[t]
    \caption{Comparison with the state-of-the-art methods on GTA5 $\rightarrow$ C-Driving. \dag~denotes methods that employ the long-training strategy.}
    \label{tab:sota-GTA}
    \small
    \centering
    \begin{tabular}{l|c|p{1cm}<{\centering}p{1cm}<{\centering}p{1cm}<{\centering}|p{1.3cm}<{\centering}|p{0.7cm}<{\centering}p{0.7cm}<{\centering}}
    \toprule
    Methods & Source & \multicolumn{3}{c|}{Compound(C)} & Open(O) & \multicolumn{2}{c}{Avg} \\
    GTA5 $\rightarrow$ & Free& Rainy & Snowy & Cloudy & Overcast & C & C+O \\
    \midrule
    Source Only & \color{ForestGreen}\cmark & 16.2 & 18.0 & 20.9 & 21.2 & 18.9 & 19.1 \\
    \midrule
    AdaptSeg~\cite{adaptseg}& \color{red}\xmark & 20.2 & 21.2 & 23.8 & 25.1 & 22.1 & 22.5 \\
    CBST~\cite{cbst}& \color{red}\xmark & 21.3 & 20.6 & 23.9 & 24.7 & 22.2 & 22.6 \\
    IBN-Net~\cite{ibn}& \color{red}\xmark & 20.6 & 21.9 & 26.1 & 25.5 & 22.8 & 23.5 \\
    PyCDA~\cite{pycda}& \color{red}\xmark & 21.7 & 22.3 & 25.9 & 25.4 & 23.3 & 23.8 \\
    Liu~\etal~\cite{liu2020open}& \color{red}\xmark & 22.0 & 22.9 & 27.0 & 27.9 & 24.5 & 25.0 \\
    Park~\etal~\cite{park2020discover}& \color{red}\xmark & 27.0 & 26.3 & 30.7 & 32.8 & 28.5 & 29.2 \\
    \midrule
    Source Only\dag & \color{ForestGreen}\cmark & 23.6 & 24.4 & 27.8 & 29.5 & 25.6 & 26.3 \\
    \midrule
    AdaptSeg~\cite{adaptseg}\dag & \color{red}\xmark & 25.6 & 27.2 & 31.8 & 32.1 & 28.8 & 29.2 \\
    MOCDA~\cite{gong2020cluster}\dag & \color{red}\xmark & 24.4 & 27.5 & 30.1 & 31.4 & 27.7 & 29.4 \\
    Park~\etal~\cite{park2020discover}\dag& \color{red}\xmark & 27.1 & 30.4 & 35.5 & 36.1 & 32.0 & 32.3 \\
    Ours~(Stage-I)\dag& \color{ForestGreen}\cmark & 28.5 & 30.5 & 36.4 & 37.4 & 32.8 & 33.2 \\
    Ours~(Stage-II)\dag & \color{ForestGreen}\cmark & \textbf{30.6} &\textbf{31.9} & \textbf{37.6} & \textbf{38.0} & \textbf{34.4} & \textbf{34.5} \\
    \bottomrule
    \end{tabular}
    \vspace{-.1in}
\end{table}

\textbf{Results of GTA5 $\rightarrow$ C-Driving.}
In Tab.~\ref{tab:sota-GTA}, we compare our method with the state-of-the-art UDA models~\cite{adaptseg, cbst, ibn, pycda} and OCDA models~\cite{liu2020open, park2020discover, gong2020cluster} on the setting of ``GTA5 to C-Driving''. For a fair comparison, all the models adopt DeepLab-V2 with VGG16 backbone.
Following~\cite{park2020discover}, we use the long training scheme~(150K iterations) to train the model. We make the following observations. First, the models trained with the long training scheme produce higher results, showing the advantage of the long training scheme. Second, our ``Stage-I'' model, which is trained only with the source data, achieves the best performance among all the existing methods that use both the source and the target data. This verifies the effectiveness of the proposed CSPP in learning a generalizable model. Third, our ``Stage-II'' model outperforms all compared models by a large margin, indicating that our method produces new state-of-the-art performance for OCDA, even under the source-free constraint.

\textbf{Results of Domain Generalization.}
We also verify the generalization ability of our method on CityScapes in Tab.~\ref{tab:cityscapes}. All models are trained with the VGG16 backbone. We can obtain the following findings. First, our ``Stage-I'' model surpasses the state-of-the-art domain generalization methods when training only with GTA5. Compared with DRPC~\cite{DRPC} that additionally uses ImageNet~\cite{imagenet} images, our model is 1.0\% higher than it. Second, when training the model with GTA5 and (unlabeled) C-Driving, our ``Stage-II'' model significantly outperforms the compared methods, MOCDA~\cite{gong2020cluster} and AdaptSeg~\cite{adaptseg} by a large margin. These two findings demonstrate the effectiveness of the proposed method on open domains.

\begin{table}[t]
\begin{minipage}{0.48\linewidth}
\caption{Evaluation on open domain CityScapes. $\ddagger$ extra using the unlabeled C-Driving. $\S$ extra using the ImageNet images.}
    \label{tab:cityscapes}
    \centering
    \small
    \begin{tabular}{l|c}
    \toprule
        Method & GTA5 $\rightarrow$ CityScapes \\
        \midrule
        ASG~\cite{ASG} & 31.5 \\
        IBN-Net~\cite{ibn} & 34.8 \\
        DRPC~\cite{DRPC}$\S$ & 36.1 \\
        Ours~(Stage-I) & \textbf{37.1} \\
        \midrule
        MOCDA~\cite{gong2020cluster}$\ddagger$ & 31.1 \\
        AdaptSeg~\cite{adaptseg}$\ddagger$ & 32.8 \\
        Ours~(Stage-II)$\ddagger$ & \textbf{38.1} \\
    \bottomrule
    \end{tabular}
\end{minipage}
\begin{minipage}{0.48\linewidth}
\centering
\small
\caption{Effectiveness of Style Augmentations.}
\label{tab:style_aug}
\begin{tabular}{l|c|c|c|c} 
\toprule  
Model & CPSS & PT & C & C+O \\
\midrule  
\multirow{3}{*}{Stage-I} & \color{red}\xmark & \color{red}\xmark & 25.6 & 26.3 \\
& \color{ForestGreen}\cmark & \color{red}\xmark  & 31.2 & 32.0 \\
& \color{ForestGreen}\cmark & \color{ForestGreen}\cmark & \textbf{32.8} & \textbf{33.2} \\
\midrule
\multirow{3}{*}{Stage-II} & \color{red}\xmark & \color{red}\xmark & 33.3 & 33.5 \\
& \color{ForestGreen}\cmark & \color{red}\xmark  & 34.3 & 34.4 \\
& \color{ForestGreen}\cmark & \color{ForestGreen}\cmark & \textbf{34.4} & \textbf{34.5} \\
\bottomrule
\end{tabular} 
\end{minipage}
\vspace{-.1in}
\end{table}

\subsection{Evaluation}
\label{sec:ablation}

\textbf{Effectiveness of Style Augmentations.} In Tab.~\ref{tab:style_aug}, we investigate the effectiveness of the proposed CPSS and photometric transformation~(PT). Clearly, CPSS consistently improves the performance for both stages. Specifically, for the source training stage~(Stage-I), inserting CPSS outperforms the baseline by 5.6\% in C mIoU and by 5.7\% in C+O mIoU. Adopting the photometric transformation further gains 1.6\% and 1.2\% improvement in C mIoU and C+O mIoU, respectively. For the target training stage~(Stage-II), we initialize the model by the source model trained with CPSS and PT. Without using style augmentations, self-supervised learning achieves limited improvement. In contrast, adding CPSS can clearly promote the performance on both compound and open domains. This verifies that CPSS can not only reduce the impact of noisy samples but also improve the robustness of the model to unseen domains. On the other hand, using photometric transformation has a slight influence on the performance. This is mainly because that the model has been familiar with such transformation during source training.

\textbf{Comparison of Different Stylized Operations.} In Tab.~\ref{tab:operations}, we compare several stylized operations that do not use any auxiliary information, \ie, MixStyle~\cite{zhou2021mixstyle}, CrossNorm~\cite{crossnorm}, and two versions of our CPSS. Results are conducted on the source training stage. We can find that mixing styles with a random weight~(MixStyle) is less suitable for semantic segmentation, because MixStyle may sometimes generate semantically unrealistic styles. Compared with CrossNorm and CPSS~(intra-image), CPSS~(inter-image) produces clearly higher performance. This indicates that augmenting samples with more various styles can help us to learn a more generalizable model in semantic segmentation.

\textbf{Is Splitting Latent Domains Necessary?} Recent OCDA methods~\cite{park2020discover, gong2020cluster} show that the sub-domain labels can be used to reduce the latent domain gaps in the target domain. Instead, in our target training stage, we randomly select training samples from the target data to form the mini-batch without considering the sub-domain labels. To verify the impact of considering the latent domains for CPSS, we implement our framework with a new sampling strategy. Specifically, we sample the images in a balanced way, so that each mini-batch contains at least one sample for each sub-domain. We provide two kinds of latent domains: ``Oracle'' denotes using the original rainy, snowy, cloudy as the latent domains; and ``Clustering'' denotes separating latent domains by clustering the style features. As shown in Tab.~\ref{tab:latent}, the random sampling strategy and its two variants achieve similar performance. This indicates that the proposed CPSS can potentially consider the style variations among different latent domains and learn a robust model, even without considering the factor of latent domains.

\begin{table}[t]
\begin{minipage}{0.55\linewidth}  
    \caption{Comparison of different stylized operations.}
    \label{tab:operations}
    \centering
    \small
\begin{tabular}{l|c|c}
    \toprule
        Method & C & C+O \\
        \midrule
        MixStyle~\cite{zhou2021mixstyle} & 30.7 & 31.2 \\
        CrossNorm~\cite{crossnorm} & 31.4 & 31.8 \\
        CPSS (intra-image) & 31.7 & 32.3 \\
        CPSS (inter-image ) & \textbf{32.8} & \textbf{33.2} \\
    \bottomrule
    \end{tabular}
\end{minipage}
\begin{minipage}{0.44\linewidth}  
\centering
\small
\caption{Impact of latent domains.}
\label{tab:latent}
\begin{tabular}{c|c|c|c} 
\toprule  
W/ Latent & Split & C & C+O \\
\midrule  
\multirow{2}{*}{\color{ForestGreen}\cmark} & Clustering & \textbf{34.4} & \textbf{34.7}  \\
& Oracle & 34.3 & 34.5 \\
\midrule
\color{red}\xmark & --- & \textbf{34.4} & 34.5 \\
\bottomrule
\end{tabular} 
\end{minipage}
\vspace{-.2in}
\end{table}

\subsection{Parameter Analysis}
We further analyze the sensitivities of CPSS to three important hyper-parameters, \ie, the activation probability $\beta$, the number of patches $n$ and the injecting location $l$. By default, we vary the value of one parameter and keep the others fixed. Experiments are conducted in the source training stage.

\begin{figure}[!t]
    \centering
    \includegraphics[width=0.95\textwidth]{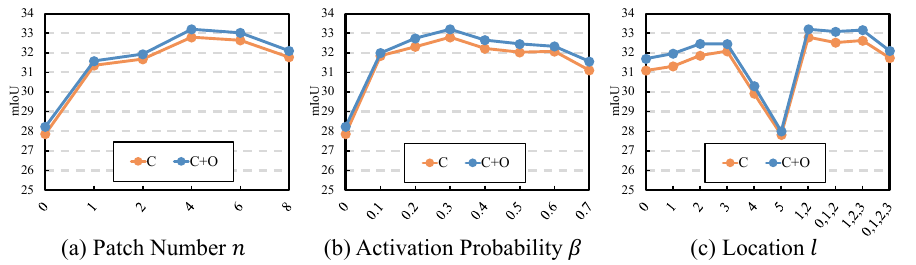}
    \vspace{-.1in}
    \caption{Sensitivities to (a) number of patches, (b) activation probability and (c) injecting location of CPSS.}
    \label{fig:parameter}
\end{figure}

\textbf{Patch Number $n$.} We compared the results of using different numbers of patches $n$ in Fig.~\ref{fig:parameter}(a). When $n$=0, the model is trained without CPSS. With the increase of $n$, the model is encouraged to face more styles, producing higher results. However, when $n$ is too large, \ie, 8, the patches are too small, which may generate less realistic styles and thus reduces the performance.

\textbf{Activation Probability $\beta$.} In Fig.~\ref{fig:parameter}(b), we investigate the effect of the probability $\beta$ of activating the CPSS operation. The performance first increases with the value of $\beta$ and achieves the best results when $\beta$=0.3. However, assigning a larger value to $\beta$ (\textit{e.g., 0.7}) leads to performance degradation. The results show that diversified styles can improve the generalization but training with excess generated styles fails to further improve the model performance.

\textbf{Injecting Location $l$.} In Fig.~\ref{fig:parameter}(c), we estimate the impact of injecting CPSS into different blocks of the network. Block-$0$ denotes injecting the CPSS before the network, and block-$l~(l>0)$ denotes injecting CPSS before the last pooling layer of the $l$th convolutional block. We make two observations. First, injecting CPSS into shallow layers, \ie, block-$0,1,2,3$, helps to improve the performance, while the performance degrades when injecting CPSS into a deep layer (block-$4,5$). The reason is that the mean and standard deviation represent style information in shallow layers but contain more semantic information in deep layers. Second, jointly injecting into multiple (two or three) layers can achieve further improvement. Considering the trade-off between accuracy and runtime, injecting CPSS into block-$1$ and block-$2$ is an appropriate choice.

\begin{figure}[t]
    \centering
    \includegraphics[width=0.95\textwidth]{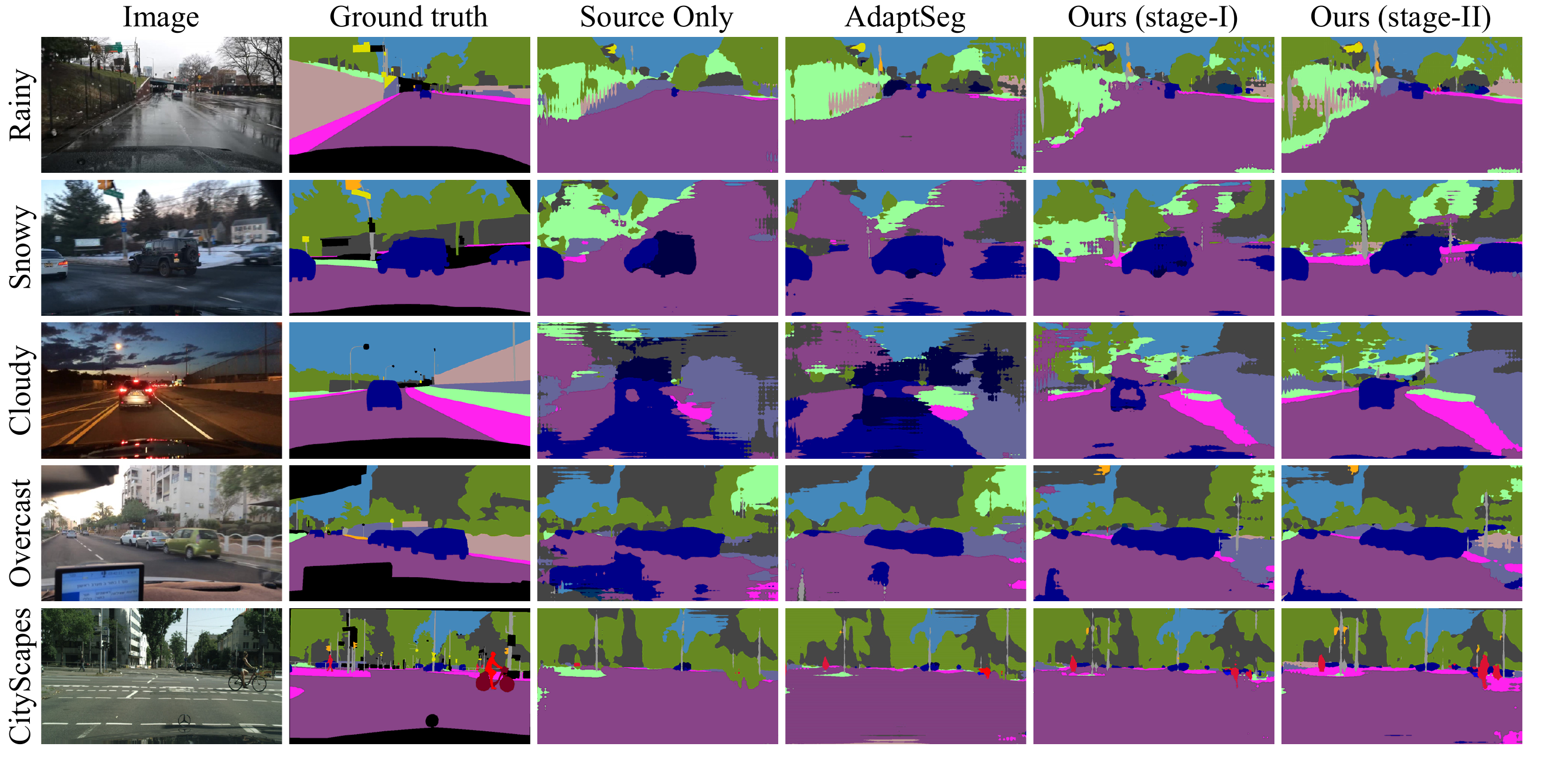}
    \vspace{-.12in}
    \caption{Qualitative comparison of segmentation result on the compound domain~(rainy, snowy, and cloudy) and open domains~(overcast and CityScapes).}
    \label{fig:seg_results}
\end{figure}

\begin{figure}[!t]
    \centering
    \vspace{-.12in}
    \includegraphics[width=0.95\textwidth]{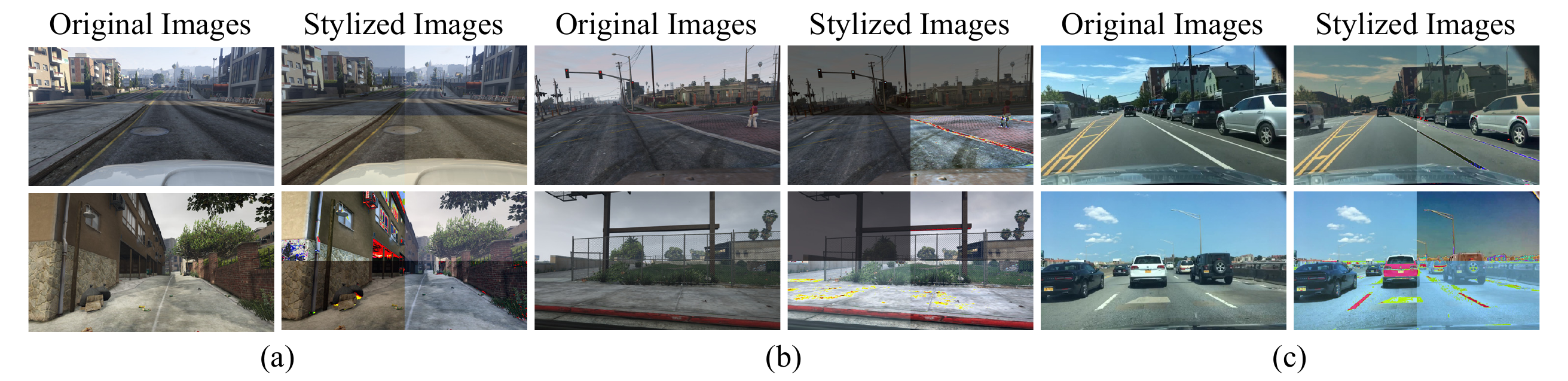}
    \vspace{-.12in}
    \caption{Examples of stylized images of CPSS. We directly apply CPSS on the image-level for image pairs on GTA5 (a and b) and C-Driving (c). The number of patches is set to 4.}
    \label{fig:cpss_visual}
    \vspace{-.2in}
\end{figure}

\subsection{Visualization}
\label{sec:visualization}
\textbf{Qualitative Comparison of Segmentation Results.} We compare the segmentation results for different models on the compound domain~(rainy, snowy, cloudy) and open domains~(overcast and CityScapes) in Fig.~\ref{fig:seg_results}. Compared with the source only model and AdaptSeg~\cite{adaptseg}, our models (Stage-I and Stage-II) clearly produce more accurate semantic results, especially for the boundaries between different objects. Comparing between our models, our Stage-II model can generate finer results on elements that have large intra-class variations between the virtual and real, \textit{e.g.}, person, car and fence.

\textbf{Image-Level Visualization of CPSS.} To better understand the effect of our CPSS in style augmentation, we visualize three groups of style exchanging in Fig.~\ref{fig:cpss_visual} by applying CPSS in the image-level (\textit{i.e.}, block-$0$). For each group, we feed two original images (left column) into CPSS and generate corresponding stylized images (right column) by swapping patch styles among the 8 (2 $\times$ 4) patches. We obverse that the styles of patches are successfully changed and various patches are generated. We can easily infer that CPSS can also change styles in the feature-level. However, the style features may contain semantic information to a certain extent, which may cause image distortion during the style exchanging. These distorted results may hammer the model optimization. For future studies, CPSS could be developed to extract more style-specific features.

\section{Conclusion}
In this work, we introduce a new setting, called source-free open compound domain adaptation (SF-OCDA), which has great potential in real-world applications. To address this problem, we propose an effective framework to train robust source and target models under the source-free constraint. Moreover, the Cross-Patch Style Swap (CPSS) is proposed to diversify the feature-level samples with various styles, consistently boosting the performance for both source and target training stages. Extensive experiments demonstrate the effectiveness of the proposed CPSS. Our method achieves state-of-the-art results on OCDA and DG benchmarks.

\appendix
\renewcommand\thesection{\Alph{section}}
\renewcommand\thefigure{\Alph{figure}} 
\renewcommand\thetable{\Alph{table}}
\setcounter{table}{0} 
\section*{Appendix}

\section{Additional Experimental Results}
\label{sec:add-experiment}
\textbf{Results of SYNTHIA $\rightarrow$ C-Driving}.
In Tab.~\ref{tab:sota-SY}, we compare our method with state-of-the-art methods on the setting of ``SYNTHIA $\rightarrow$ C-Driving''. Clearly, (1) the proposed method largely improves the performance of the source only model, and (2) our two models (Stage-I and Stage-II) both significantly outperform the state-of-the-art methods, verifying the generalization ability of the proposed method with different source datasets. 
We also find that the improvement of our Stage-II is limited. This is because given a poorly trained source model ($\approx$ 24\% mIoU), we fail to generate enough useful / reliable pseudo-labels for self-supervised learning on the target domain. In our experiments, training the target model without the proposed CPSS will reduce the performance. 
This phenomenon can also be observed for Advent~\cite{advent}, which additional uses entropy information to train the AdaptSeg~\cite{adaptseg} but achieves lower results on C-Driving (in Tab.~\ref{tab:sota-SY}). In contrast, using our CPSS can alleviate the impact of wrong pseudo-labels and can guarantee that self-supervised learning will not hamper the model performance.

\begin{table}[ht]
    \caption{Comparison with the state-of-the-art methods on SYNTHIA $\rightarrow$ C-Driving. All models are trained with the long-training strategy. We report averaged performance on 16 class subsets following the evaluation protocol used in \cite{advent,park2020discover}. $^\ast$ denotes the source only model trained in this paper.}
    \label{tab:sota-SY}
    \centering
    \begin{tabular}{l|c|p{1cm}<{\centering}p{1cm}<{\centering}p{1cm}<{\centering}|p{1.3cm}<{\centering}|p{0.7cm}<{\centering}p{0.7cm}<{\centering}}
    \toprule
    Methods & Source & \multicolumn{3}{c|}{Compound(C)} & Open(O) & \multicolumn{2}{c}{Avg} \\
    SYNTHIA $\rightarrow$& Free& Rainy & Snowy & Cloudy & Overcast & C & C+O \\
    \midrule
    Source Only~\cite{park2020discover}& \color{ForestGreen}\cmark & 16.3 &18.8 &19.4 &19.5 &18.4 &18.5 \\
    \midrule
    CBST~\cite{cbst}& \color{red}\xmark & 16.2 & 19.6 & 20.1 & 20.3 & 18.9 & 19.1 \\
    CRST~\cite{crst}& \color{red}\xmark & 16.3 & 19.9 & 20.3 & 20.5 & 19.1 & 19.3 \\
    AdaptSeg~\cite{adaptseg}& \color{red}\xmark & 17.0 & 20.5 & 21.6 & 21.6 & 20.0 & 20.2 \\
    Advent~\cite{advent}& \color{red}\xmark & 17.7 & 19.9 & 20.2 & 20.5 & 19.3 & 19.6 \\
    Park~\etal~\cite{park2020discover}& \color{red}\xmark & 18.8 & 21.2 & 23.6 & 23.6 & 21.5 & 21.8\\
    \midrule
    Source Only$^\ast$& \color{ForestGreen}\cmark & 18.9 & 19.7 & 20.4 & 21.3 & 19.7 & 20.1 \\
    Ours~(Stage-I)& \color{ForestGreen}\cmark & \textbf{22.4}& 23.8 & \textbf{25.3} & \textbf{26.4} & 24.0 & 24.5 \\
    Ours~(Stage-II)& \color{ForestGreen}\cmark & \textbf{22.4}& \textbf{24.5} & \textbf{25.3} & \textbf{26.4} & \textbf{24.2} & \textbf{24.7}  \\
    \bottomrule
    \end{tabular}
\end{table}

\textbf{Implementing AdaptSeg with CPSS}.
To further demonstrate the generalization ability of the proposed CPSS, we inject CPSS into the widely used domain adaptation approach, AdaptSeg~\cite{adaptseg}, and evaluate the results on the settings of ``GTA5 $\rightarrow$ C-Driving'' and ``GTA5 $\rightarrow$ CityScapes''. Note that, when using AdaptSeg, the source-free constraint is not enforced. Clearly, CPSS can consistently improve the performance of AdaptSeg by a large margin on both settings. This further confirms the compatibility of the proposed CPSS.

\begin{table}[ht]
    \caption{Effectiveness of CPSS in AdaptSeg model for OCDA (GTA5$\rightarrow$C-Driving) and UDA (GTA5$\rightarrow$CityScapes). $^\ast$ denotes reproducing the method based on the source code.}
    \label{tab:cpss_adaptseg}
    \centering
    \small
    \begin{tabular}{l|c|p{0.8cm}<{\centering}p{0.8cm}<{\centering}p{0.8cm}<{\centering}|p{1.3cm}<{\centering}|p{0.5cm}<{\centering}p{0.5cm}<{\centering}|c}
    \toprule
    \multirow{3}{*}{Methods} & \multirow{3}{*}{CPSS} & \multicolumn{6}{c|}{GTA5$\rightarrow$C-Driving} & \multirow{3}{*}{GTA5$\rightarrow$CityScapes} \\
    \cline{3-8}
    & & \multicolumn{3}{c|}{Compound(C)} & Open(O) & \multicolumn{2}{c|}{Avg} & \\
    & &Rainy & Snowy & Cloudy & Overcast & C & C+O & \\
    \midrule
    AdaptSeg~\cite{adaptseg} & \color{red}\xmark & --- & --- & --- & --- & --- & --- & 35.0 \\
    AdaptSeg~\cite{adaptseg}$^\ast$ & \color{red}\xmark & 25.6 & 27.2 & 31.8 & 32.1 & 28.8 & 29.2 & 34.2 \\
    AdaptSeg~\cite{adaptseg} & \color{ForestGreen}\cmark & \textbf{28.9} & \textbf{29.1} & \textbf{35.2} & \textbf{36.0} & \textbf{31.9} & \textbf{32.3} & \textbf{38.5} \\
    \bottomrule
    \end{tabular}
\end{table}

\textbf{Per-Class IoU on GTA5 $\rightarrow$ C-Driving}.
In Tab.~\ref{tab:gtaperclass_target}, we report the per-class IoU on different sub-domains of ``GTA5$\rightarrow$C-Driving''. Generally, our methods (Stage-I and Stage-II) produce higher results on most classes for all sub-domains. On the other hand, we find that all the methods fail to recognize the samples of the ``train'', ``motorcycle'' and ``“bicycle'' classes, which are rarely appeared in the C-Driving dataset.

\begin{table}[!ht]
    \centering
    \caption{Per-Class IoU on different sub-domains of the OCDA benchmark: GTA5 $\rightarrow$ C-Driving. The rainy, snowy and cloudy weather compose the compound target domain, while the overcast weather is the open domain. The results are reported over 19 classes. The ``bicycle'' class is not listed due to the result is close to zero. The best results are denoted in bold. \dag~denotes methods that employ the long-training strategy.}
    \label{tab:gtaperclass_target}
    \resizebox{\textwidth}{!}{
    \begin{tabular}{c|c|cccccccccccccccccc|c}
    \toprule
    \multicolumn{21}{c}{GTA5$\rightarrow$C-Driving}\\
    \hline
    Sub-domain & Method &\rotatebox{90}{road}&\rotatebox{90}{sidewalk}&\rotatebox{90}{building}&\rotatebox{90}{wall}&\rotatebox{90}{fence}&\rotatebox{90}{pole}&\rotatebox{90}{light}&\rotatebox{90}{sign}&\rotatebox{90}{vegetation}&\rotatebox{90}{terrain}&\rotatebox{90}{sky}&\rotatebox{90}{person}&\rotatebox{90}{rider}&\rotatebox{90}{car}&\rotatebox{90}{truck}&\rotatebox{90}{bus}&\rotatebox{90}{train}&\rotatebox{90}{motocycle}&mIoU\\
    \hline
    \multirow{10}{*}{Rainy} & Source Only~\cite{liu2020open} & 48.3 & 3.4 & 39.7 & 0.6 & 12.2 & 10.1 & 5.6 & 5.1 & 44.3 & 17.4 & 65.4 & 12.1 & 0.4 & 34.5 & 7.2 & 0.1 & 0.0 & 0.5 & 16.2 \\
    & AdaptSegNet~\cite{adaptseg,liu2020open}& 58.6 & 17.8 & 46.4 & 2.1 & 19.6 & 15.6 & 5.0 & 7.7 & 55.6 & \textbf{20.7} & 65.9 & 17.3 & 0.0 & 41.3 & 7.4 & 3.1 & 0.0 & 0.0 & 20.2 \\
    & CBST~\cite{cbst,liu2020open} & 59.4 & 13.2 & 47.2 & 2.4 & 12.1 & 14.1 & 3.5 & 8.6 & 53.8 & 13.1 & 80.3 & 13.7 & \textbf{17.2} & 49.9 & 8.9 & 0.0 & 0.0 & \textbf{6.6} & 21.3 \\
    & IBN-Net~\cite{ibn,liu2020open} & 58.1 & 19.5 & 51.0 & 4.3 & 16.9 & 18.8 & 4.6 & 9.2 & 44.5 & 11.0 & 69.9 & 20.0 & 0.0 & 39.9 & 8.4 & 15.3 & 0.0 & 0.0 & 20.6 \\
    & OCDA~\cite{liu2020open} & 63.0 & 15.4 & 54.2 & 2.5 & 16.1 & 16.0 & 5.6 & 5.2 & 54.1 & 14.9 & 75.2 & 18.5 & 0.0 & 43.2 & 9.4 & 24.6 & 0.0 & 0.0 & 22.0 \\
    & MOCDA~\cite{gong2020cluster}\dag & 66.8 & 22.0 & 52.4 & 6.7 & 16.7 & 16.9 & 5.3 & 3.5 & 60.4 & 17.2 & 80.1 & 21.8 & 0.1 & 46.4 & 17.9 & \textbf{29.4} & 0.0 & 0.0 & 24.4\\
    & Source Only\dag & 65.8 & 17.2 & 59.8 & 7.0  & 8.5  & 15.6 & 3.1  & 5.6  & 59.9 & 13.8 & \textbf{80.8} & 21.4 & 0.0  & 47.3 & 23.3 & 18.5 & 0.0 & 0.0  & 23.6 \\
    & AdaptSeg~\cite{adaptseg}\dag & 63.9 & 17.9 & 60.7 & 9.6  & 15.0 & 16.8 & 6.5  & 11.5 & 61.2 & 15.3 & 78.5 & 24.4 & 14.4 & 53.4 & 18.3 & 14.5 & 0.0 & 3.6  & 25.6 \\
    & Ours~(Stage-I)\dag & 75.0 & 31.5 & 65.0 & 11.3 & 19.5 & 22.0 & 8.6  & 14.7 & 61.3 & 17.9 & 79.3 & \textbf{29.6} & 3.0  & 64.1 & 20.7 & 16.9 & 0.0 & 0.3  & 28.5 \\
    & Ours~(Stage-II)\dag & \textbf{78.5} & \textbf{36.6} & \textbf{65.7} & \textbf{12.9} & \textbf{23.9} & \textbf{25.4} & \textbf{9.8}  & \textbf{16.3} & \textbf{62.6} & 16.8 & 80.7 & 29.1 & 0.0  & \textbf{67.5} & \textbf{30.1} & 23.2 & 0.0 & 1.7  & \textbf{30.6} \\
    
    \hline
     \multirow{10}{*}{Snowy} & Source Only~\cite{liu2020open} & 50.8 & 4.7 & 45.1 & 5.9 & \textbf{24.0} & 8.5 & 10.8 & 8.7 & 35.9 & 9.4 & 60.5 & 17.3 & 0.0 & 47.7 & 9.7 & 3.2 & 0.0 & 0.7 & 18.0\\
    & AdaptSegNet~\cite{adaptseg,liu2020open}& 59.9 & 13.3 & 52.7 & 3.4 & 15.9 & 14.2 & 12.2 & 7.2 & 51.0 & \textbf{10.8} & 72.3 & 21.9 & 0.0 & 55.0 & 11.3 & 1.7 & 0.0 & 0.0 & 21.2\\
    & CBST~\cite{cbst, liu2020open} & 59.6 & 11.8 & 57.2 & 2.5 & 19.3 & 13.3 & 7.0 & 9.6 & 41.9 & 7.3 & 70.5 & 18.5 & 0.0 & 61.7 & 8.7 & 1.8 & 0.0 & 0.2 & 20.6\\
    & IBN-Net~\cite{ibn, liu2020open} & 61.3 & 13.5 & 57.6 & 3.3 & 14.8 & 17.7 & 10.9 & 6.8 & 39.0 & 6.9 & 71.6 & 22.6 & 0.0 & 56.1 & 13.8 & 20.4 & 0.0 & 0.0 & 21.9\\
    & OCDA~\cite{liu2020open} & 68.0 & 10.9 & 61.0 & 2.3 & 23.4 & 15.8 & 12.3 & 6.9 & 48.1 & 9.9 & 74.3 & 19.5 & 0.0 & 58.7 & 10.0 & 13.8 & 0.0 & 0.1 & 22.9\\
    & MOCDA~\cite{gong2020cluster}\dag & 71.8 & 16.9 & 61.1 & 6.5 & 21.4 & 16.3 & 17.0 & 7.5 & 52.9 & 8.7 & 79.7 & 29.2 & 0.5 & 62.7 & 18.9 & \textbf{29.4} & 0.0 & \textbf{22.6} & 27.5 \\
    & Source Only\dag & 68.1 & 11.7 & 65.5 & 7.9  & 16.0 & 16.3 & 10.0 & 5.1  & \textbf{55.0} & 5.9  & \textbf{81.6} & 27.4 & 0.0  & 63.5 & 18.8 & 10.6 & 0.0 & 0.0  & 24.4 \\
    & AdaptSeg~\cite{adaptseg}\dag & 65.3 & 12.6 & 68.6 & 15.6 & 19.8 & 17.6 & 17.7 & 11.6 & 51.0 & 6.8  & 79.3 & 35.3 & \textbf{6.5}  & 63.5 & 15.7 & 21.2 & 0.0 & 9.4  & 27.2 \\
    & Ours~(Stage-I)\dag & 81.8 & 20.0 & 70.8 & 19.6 & 20.8 & 18.9 & 21.4 & 15.4 & 52.1 & 8.5  & 78.6 & 36.0 & 0.6  & 74.4 & 25.9 & 20.2 & 0.0 & 14.7 & 30.5 \\
    & Ours~(Stage-II)\dag & \textbf{83.4} & \textbf{22.7} & \textbf{71.6} & \textbf{21.3} & 21.9 & \textbf{21.9} & \textbf{23.1} & \textbf{17.6} & 54.2 & 9.2  & 80.8 & \textbf{36.8} & 0.0  & \textbf{74.7} & \textbf{29.8} & 28.9 & 0.0 & 15.9 & \textbf{31.9} \\
    
    \hline
    \multirow{10}{*}{Cloudy} & Source Only~\cite{liu2020open} & 47.0 & 8.8 & 33.6 & 4.5 & 20.6 & 11.4 & 13.5 & 8.8 & 55.4 & 25.2 & 78.9 & 20.3 & 0.0 & 53.3 & 10.7 & 4.6 & 0.0 & 0.0 & 20.9\\
    & AdaptSegNet~\cite{adaptseg, liu2020open}& 51.8 & 15.7 & 46.0 & 5.4 & 25.8 & 18.0 & 12.0 & 6.4 & 64.4 & 26.4 & 82.9 & 24.9 & 0.0 & 58.4 & 10.5 & 4.4 & 0.0 & 0.0 & 23.8\\
    & CBST~\cite{cbst, liu2020open} & 56.8 & 21.5 & 45.9 & 5.7 &  19.5 & 17.2 & 10.3 & 8.6 & 62.2 & 24.3 & 89.4 & 20.0 & 0.0 & 58.0 & 14.6 & 0.1 & 0.0 & 0.1 & 23.9\\
    & IBN-Net~\cite{ibn, liu2020open} & 60.8 & 18.1 & 50.5 & 8.2 & 25.6 & 20.4 & 12.0 & 11.3 & 59.3 & 24.7 & 84.8 & 24.1 & 12.1 & 59.3 & 13.7 & 9.0 & 0.0 & 1.2 & 26.1\\
    & OCDA~\cite{liu2020open} & 69.3 & 20.1 & 55.3 & 7.3 & 24.2 & 18.3 & 12.0 & 7.9 & 64.2 & 27.4 & 88.2 & 24.7 & 0.0 & 62.8 & 13.6 & 18.2 & 0.0 & 0.0 & 27.0\\
    & MOCDA~\cite{gong2020cluster}\dag & 79.6 & 21.7 & 61.4 & 11.0 & 27.6 & 19.4 & 13.4 & 8.3 & 69.0 & 26.4 & 89.1 & 25.0 & 3.2 & 69.5 & 22.7 & \textbf{21.5} & 0.0 & 3.5 & 30.1\\
    & Source Only\dag & 70.1 & 16.0 & 64.1 & 8.5  & 26.9 & 17.6 & 9.3  & 7.6  & 69.5 & 23.5 & 87.0 & 25.7 & 0.0  & 66.1 & 26.6 & 8.9  & 0.0 & 0.0  & 27.8 \\
    & AdaptSeg~\cite{adaptseg}\dag & 69.1 & 21.0 & 67.2 & 12.9 & 35.2 & 20.0 & 14.8 & 17.1 & 72.7 & 24.2 & 88.7 & 32.9 & 23.1 & 58.6 & 26.5 & 14.3 & 0.0 & 5.5  & 31.8 \\
    & Ours~(Stage-I)\dag & 85.2 & 30.9 & 69.1 & 20.3 & 34.6 & 21.4 & 15.9 & 20.4 & 72.8 & 30.4 & 88.9 & \textbf{38.8} & 32.4 & 77.3 & 33.6 & 8.4  & 0.0 & 11.4 & 36.4 \\
    & Ours~(Stage-II)\dag & \textbf{86.1} & \textbf{35.7} & \textbf{69.9} & \textbf{21.3} & \textbf{36.9} & \textbf{24.5} & \textbf{16.9} & \textbf{23.0} & \textbf{73.7} & \textbf{31.0} & \textbf{89.9} & 37.0 & \textbf{33.1} & \textbf{78.0} & \textbf{36.5} & 10.2 & 0.0 & \textbf{11.6} & \textbf{37.6} \\
    
    \hline
    \multirow{10}{*}{Overcast} & Source Only~\cite{liu2020open} & 46.6 & 9.5 & 38.5 &  2.7 & 19.8  & 12.9 & 9.2 & 17.5 &  52.7 &  19.9 & 76.8 &  20.9 & 1.4 & 53.8 & 10.8 & 8.4 & 0.0 & 1.8 & 21.2\\
    & AdaptSegNet~\cite{adaptseg, liu2020open}& 59.5 & 24.0 & 49.4 &  6.3 & 23.3 & 19.8 & 8.0 & 14.4 & 61.5 & 22.9 & 74.8 & 29.9 & 0.3 & 59.8 &  12.8 & 9.7 & 0.0 & 0.0 & 25.1 \\
    & CBST~\cite{cbst, liu2020open} & 58.9 & 26.8 & 51.6 & 6.5 & 17.8 & 17.9 & 5.9 & 17.9 & 60.9 & 21.7 & 87.9 & 22.9 & 0.0 & 59.9 & 11.0 & 2.1 & 0.0 & 0.2 & 24.7\\
    & IBN-Net~\cite{ibn, liu2020open} & 62.9 & 25.3 & 55.5 & 6.5 &  21.2 & 22.3 & 7.2 & 15.3 & 53.3 & 16.5 & 81.6 & 31.1 & 2.4 & 59.1 & 10.3 & 14.2 & 0.0 & 0.0 & 25.5\\
    & OCDA~\cite{liu2020open} & 73.5 & 26.5 & 62.5 & 8.6 & 24.2 & 20.2 & 8.5 & 15.2 & 61.2 & 23.0 & 86.3 & 27.3 & 0.0 & 64.4 & 14.3 & 13.3 & 0.0 & 0.0 & 27.9\\
    & MOCDA~\cite{gong2020cluster}\dag & 80.1 & 28.6 & 66.0 & 13.0 & 26.6 & 20.9 & 8.9 & 15.5 & 67.0 & 25.1 & 87.7 & 33.2 & 9.5 & 69.2 & 23.0 & 18.3 & \textbf{2.2} & 2.0 & 31.4\\
    & Source Only\dag & 72.9 & 23.3 & 68.8 & 10.1 & 19.7 & 18.8 & 6.2  & 11.3 & 69.0 & 23.1 & 87.5 & 36.1 & 10.5 & 67.8 & 26.3 & 9.4  & 0.0 & 0.0  & 29.5 \\
    & AdaptSeg~\cite{adaptseg}\dag & 69.9 & 26.4 & 71.0 & 14.9 & 25.6 & 21.1 & 11.5 & 22.1 & 70.0 & 25.5 & 87.9 & 39.6 & 20.8 & 61.7 & 25.2 & 13.9 & 0.0 & 2.0  & 32.1 \\
    & Ours~(Stage-I)\dag & 85.1 & 38.3 & 73.5 & 25.3 & 29.0 & 24.5 & 12.4 & 26.2 & 70.9 & \textbf{32.1} & 88.3 & \textbf{46.1} & \textbf{22.5} & \textbf{76.0} & 31.0 & \textbf{21.7} & 0.7 & 7.2  & 37.4 \\
    & Ours~(Stage-II)\dag & \textbf{86.0} & \textbf{41.2} & \textbf{73.9} & \textbf{25.7} & \textbf{30.6} & \textbf{27.7} & \textbf{13.6} & \textbf{27.4} & \textbf{71.9} & 31.8 & \textbf{89.3} & 44.3 & 17.5 & 75.9 & \textbf{37.0} & 21.6 & 0.0 & \textbf{7.4}  & \textbf{38.0} \\
    \bottomrule
    \end{tabular}
    }
\end{table}

\section{Comparison of Different Stylized Operations}
\label{sec:operations}

The proposed Cross-Patch Style Swap (CPSS) is closely related to MixStyle~\cite{zhou2021mixstyle} and CrossNorm~\cite{crossnorm}, which are both designed for domain generalization. All three methods aim to improve the generalization ability of the model by perturbing style features of training samples. However, the stylized operations of them are different. Specifically, MixStyle replaces the style of a sample with the one that is generated by mixing its own style feature with a shuffled style feature using a random convex weight. Instead, CrossNorm directly exchanges the styles of two samples, which is a special case of MixStyle when the weight of the shuffled style feature is 1. Both MixStyle and CrossNorm compute one style feature for each sample and stylize each sample with one style feature. Different from them, our CPSS generates several styles for each sample by separating the feature map into different patches. This modification is specially designed for semantic segmentation in the self-driving scenario, because patches in a frame could contain different styles. 
Compared with MixStyle and CrossNorm, our CPSS can provide more diverse and useful styles for generating stylized feature maps. In addition, with CPSS, the model is trained with richer feature maps where each one jointly contains multiple different styles, further enforcing the model to be robust to style variations. In our main paper, we conduct experiments by comparing these three methods and show that CPSS produces clearly higher results than the other two methods.

\small
\bibliography{references.bib}
\bibliographystyle{plainnat}


\end{document}